\documentclass[letterpaper,journal]{IEEEtran}
\usepackage{amsmath,amsfonts}
\usepackage{algorithmic}
\usepackage{algorithm}
\usepackage{array}
\usepackage[caption=false,font=normalsize,labelfont=sf,textfont=sf]{subfig}
\usepackage{textcomp}
\usepackage{stfloats}
\usepackage{url}
\usepackage{verbatim}
\usepackage{graphicx}
\usepackage{cite}

\usepackage{microtype}
\usepackage{graphicx}
\usepackage{subcaption}
\usepackage{booktabs}
\usepackage{algorithm} 
\usepackage{algorithmic}

\usepackage{multirow}
\usepackage{float}
\usepackage{siunitx}
\usepackage{silence}
\usepackage{stfloats}
\usepackage[table]{xcolor}
\usepackage[textsize=tiny]{todonotes}
\usepackage{hyperref}
\hypersetup{
    colorlinks=true,
    linkcolor=black,
    citecolor=black,
    urlcolor=black
}
\usepackage[capitalize,noabbrev]{cleveref}
\usepackage{tabularx}
\graphicspath{{figures/}{../figures/}}

\begin{document}

\title{Read-Best Is Not Steer-Best: A Probing--Steering Layer Dissociation in Omni-Modal Large Language Models}

\author{%
Yibo~Wang,
Jisheng~Dang,
Bimei~Wang,
Yitao~Wu,
Wencan~Zhang,
Hong~Peng,
Jizhao~Liu,
Bin~Hu,~\IEEEmembership{Fellow,~IEEE,} Qi Tian, ~\IEEEmembership{Fellow, ~IEEE},
and~Tat-Seng~Chua%
\thanks{Yibo Wang, Jisheng Dang, Bimei Wang, Hong Peng, Jizhao Liu, and Bin Hu are with Lanzhou University, Lanzhou, China.}%
\thanks{Yitao Wu is with Hainan University, Haikou, China.}%
\thanks{Qi Tian is with Cloud and AI BU, Huawei, Shenzhen, Guangdong 518129,
China (e-mail: tian.qi1@huawei.com).}
\thanks{Wencan Zhang and Tat-Seng Chua are with the School of Computing, National University of Singapore, Singapore.}%
\thanks{Corresponding authors: Bin Hu, Jisheng Dang, and Hong Peng.}%
}



\maketitle

\begin{abstract}
Omni-modal large language models fold text, audio, and image signals into a single residual stream, where the emotion carried by an image or a voice can be linearly read out and causally rewritten by activation steering. Practice, however, rests on a rarely tested assumption: the layer where a linear probe reads a concept most strongly is also the layer where injecting that concept steers behavior most effectively, so injection layers are chosen by probing accuracy. We give the first causal test of this assumption across three independently built omni-modal models. The assumption fails. Reading and intervention are distinct operations on the residual stream, carried by different layers, a phenomenon we call the probing--steering layer dissociation. We use emotion as a controlled vehicle. It is linearly readable across the three modalities and steerable by directional injection, so readability and steerability become two layer-wise curves in one representation space. The probe-best layer scatters across nearly the full network depth and shifts with architecture, whereas the steering-effective layer does not. It is an architectural invariant, landing in the same narrow mid-to-late band of normalized depth in every model. With paired random-direction controls, the causal gap between the two is roughly twenty-six-fold, ruling out direction quality and random perturbation. A logit-lens analysis ties the dissociation to a staged forward pass (a causal-handle, probing-saturation, vocabulary-commitment ordering) and motivates a two-factor account that links steering effectiveness to representational readability and downstream plasticity. Choosing the injection layer by probing accuracy is a mistaken heuristic, and the mid-to-late band gives a cross-architecture selection criterion. As a by-product, we map a cross-modal emotion subspace organized by valence--arousal and anchored across models by joy. Code and data are available at \url{https://github.com/YiboWang2002/Read-Best-Is-Not-Steer-Best}.
\end{abstract}


\section{Introduction}

\IEEEPARstart{O}{mni}-modal large language models process text, audio, and image signals in one shared residual stream, and two operations dominate how we read and change the concepts these signals carry. A linear probe~\cite{alain2016probes} \emph{reads} the concept encoded in a layer's activations, while activation steering~\cite{zou2023representation,turner2023steering} \emph{edits} those activations to change behavior. In practice the two are tied together by an assumption that is rarely stated and almost never tested: the layer where a probe reads a concept most strongly is taken to be the layer where injecting it steers behavior most effectively. Injection layers are then chosen directly by probing accuracy. If the assumption is wrong, much of the steering literature has been intervening at causally near-inert layers, and underestimating what these methods can do.

\begin{figure}[t]
\centering
\includegraphics[width=1\linewidth]{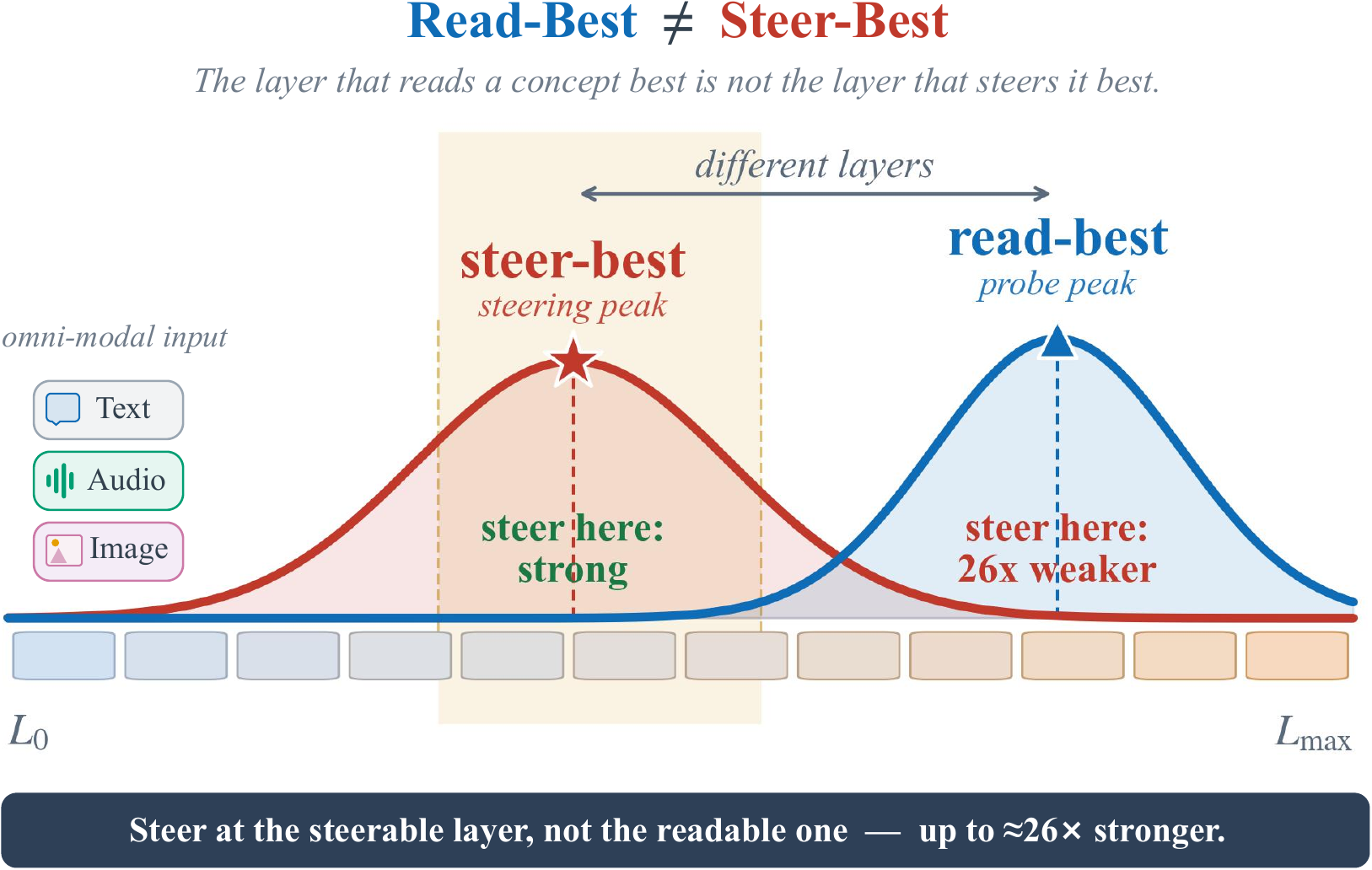}
\caption{Read-best is not steer-best. In an omni-modal LLM, probe readability (blue) peaks at a late layer while the steering effect of an emotion direction (red) peaks earlier, in the mid-to-late zone. Injecting at the read-best layer is about $26\times$ weaker than in the steering zone. Curves are schematic. Full results in \Cref{fig:teaser} and \S5.}
\label{fig:motivation}
\end{figure}

We show that this assumption fails in omni-modal language models, and we measure how far. Reading and intervention are distinct operations carried by different layers of the residual stream (\Cref{fig:motivation}). The separation is systematic, and the layer-selection heuristic at the heart of representation engineering does not transfer to these models.

\begin{figure*}[t]
\centering
\includegraphics[width=1\textwidth]{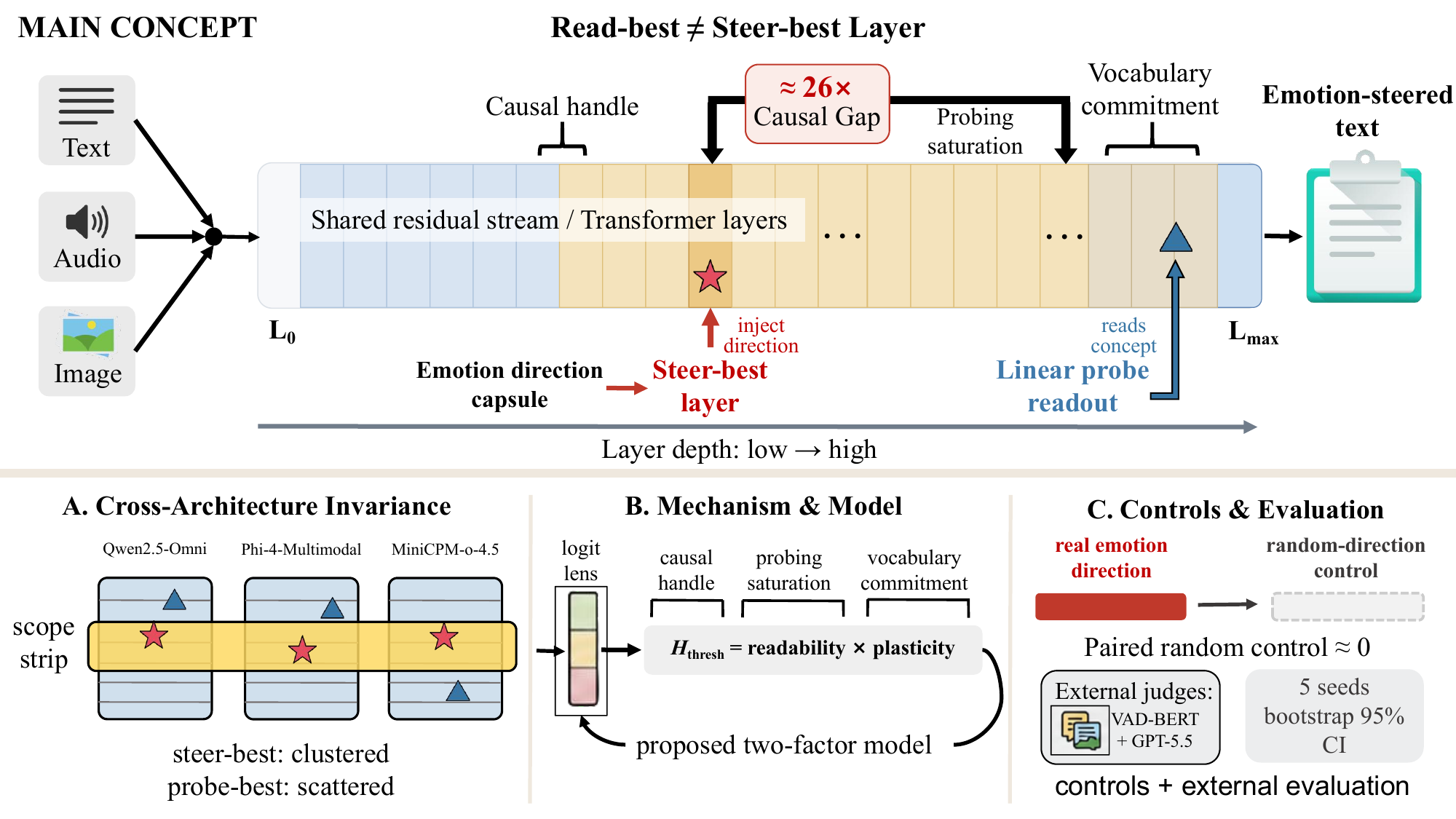}
\caption{Overview of the study. Text, audio, and image inputs enter an omni-modal LLM through a shared residual stream. A linear probe reads a concept best at a late layer (probe-best), while steering an emotion direction is most effective at a mid-to-late layer (steer-best), a controlled gap of about $26\times$. A logit-lens analysis orders the forward pass as causal handle, probing saturation, then vocabulary commitment, placing the steering window before commitment and motivating the two-factor model $H_{\mathrm{thresh}}$. Bottom: (A) the steer-best layer clusters in a shared band across three architectures while the probe-best layer scatters; (B) the logit-lens mechanism; (C) controls and external judges (VAD-BERT, GPT-5.5).}
\label{fig:overview}
\end{figure*}

\begin{figure*}[t]
\centering
\includegraphics[width=0.9\textwidth]{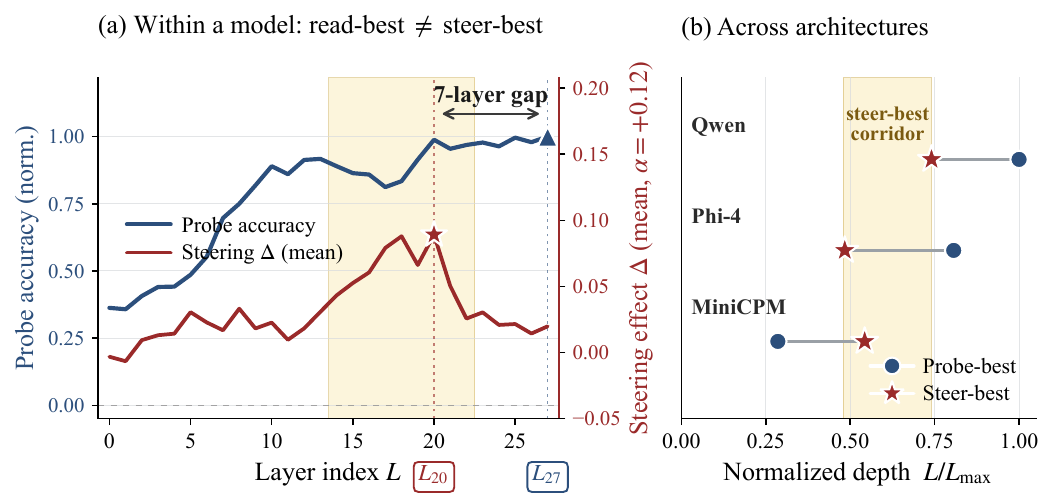}
\caption{The probing--steering layer dissociation at a glance. \textbf{(a)} On Qwen2.5-Omni-7B the probe-accuracy peak (L27) and the steering-effect peak (L20) are 7 layers apart, with the band marking the mid-to-late steering zone [L14, L22]. \textbf{(b)} Across three architectures the steer-best layer stays in a narrow band of normalized depth $[0.48, 0.74]$, while the probe-best layer scatters across $[0.29, 1.00]$ (MiniCPM shows a reversed ordering). Full evidence in \S5--\S6.}
\label{fig:teaser}
\end{figure*}

We study this probing--steering layer dissociation using emotion as a controlled vehicle, across three independently designed omni-modal LLMs (Qwen2.5-Omni-7B, Phi-4-Multimodal, MiniCPM-o-4.5). Emotion is well suited to the task: it is shared across text, audio, and image, reads out linearly, and responds to directional injection, so we can measure readability and steerability as two layer-wise curves in the same representation space and compare where they peak. \Cref{fig:overview} gives an overview of the study, from the omni-modal setup to the cross-architecture findings, mechanism, and controls.

The picture is consistent across the three architectures: the probe-best layer scatters across nearly the full network depth while the steer-best layer settles into a narrow mid-to-late band (\Cref{fig:teaser}), and on Qwen2.5-Omni single-layer injection at the probe-best layer is indistinguishable from a random direction. A logit-lens analysis locates the cause in the staged forward pass, in the window where a representation is already stable but not yet committed to a token.

We make four contributions:
\begin{itemize}
\item We provide the first causal characterization of the probing--steering dissociation in the omni-modal setting, across three independently designed omni-modal LLMs spanning text, audio, and image. The steering-effective zone is an architectural invariant. It stays in the same mid-to-late band in every model, whereas the probe-best layer varies across almost the full network depth. We frame this stability as an architecture-modulated regularity rather than a universal law (\S7.2).
\item We quantify the dissociation with a controlled causal gap of roughly 26$\times$, using paired random-direction controls to rule out direction quality and random perturbation.
\item We trace the dissociation to a logit-lens forward-pass account (a causal handle, then probing saturation, then vocabulary commitment) and propose a two-factor account, $H_{\mathrm{thresh}}$, that relates steering effectiveness to representational readability and downstream plasticity, as a first step toward predicting the intervention-optimal layer (\S5.5).
\item We draw the direct consequence for representation engineering. Choosing the injection layer by probing accuracy is a mistaken heuristic, and the mid-to-late zone gives a cross-architecture selection criterion. As a substrate for these mechanisms, we also characterize a cross-modal emotion geometry, a shared subspace organized by valence--arousal and anchored across models by joy.
\end{itemize}

\S3 describes the models, data, and protocol. \S4 characterizes the cross-modal emotion geometry. \S5 presents the core dissociation evidence. \S6 extends it to tri-modal, cross-model causal control. \S7 discusses the implications and the geometric basis.

\section{Related Work}

\subsection{Representation Engineering and Activation Steering}

Representation engineering, introduced by Zou et al.~\cite{zou2023representation}, treats high-level concepts as linear directions in activation space and intervenes on behavior by editing the residual stream. A family of methods in this vein extracts concept directions from contrastive activations and injects them additively, covering truthfulness, general behavior steering, and style or emotion control~\cite{li2023inference,turner2023steering,rimsky2024steering,konen2024style}, with roots in controllable text generation~\cite{dathathri2020pplm,keskar2019ctrl,krause2021gedi} and extensions to function vectors, in-context vectors, and refusal or conditional steering~\cite{todd2024function,liu2024incontext,arditi2024refusal,lee2025programming}. Almost all of these treat steering as \emph{single-layer} injection with the layer chosen by probing signal or empirical search, tacitly assuming that where a concept can be read is where it can be intervened upon. Closest to our work, Tan et al.~\cite{tan2024analysing} note that steering is highly sensitive to the injection layer and setup, but do not characterize the systematic read-best/steer-best separation as a phenomenon in its own right; that separation is the entry point of this work.

\subsection{Mechanistic Interpretability of Layer-Wise Function}

Mechanistic-interpretability research has characterized how information forms progressively along a transformer's layers. The logit lens~\cite{nostalgebraist2020logitlens} projects each layer's hidden state onto the vocabulary and the tuned lens~\cite{belrose2023tunedlens} calibrates this readout, while linear probes are the standard tool for reading intermediate layers~\cite{alain2016probes,hewitt2019structural,belinkov2022probing}. At the circuit level, FFN key-value memories~\cite{geva2021keyvalue,geva2022promoting}, factual-association editing~\cite{meng2022rome}, circuit analyses~\cite{wang2022ioi,conmy2023acdc}, and sparse-autoencoder studies support a picture in which concepts exist as decomposable directions, aligning with the linear representation hypothesis~\cite{park2024linear,park2024geometry} and superposition~\cite{elhage2022superposition}. These works focus on where information is \emph{represented}; whether that locus coincides with the locus of causal \emph{intervenability} is exactly what our probing--steering dissociation characterizes.

\subsection{Emotion Representation in Multimodal LLMs}

Omni-modal LLMs process text, audio, and image in a unified residual stream. The three models we study, Qwen2.5-Omni, Phi-4-Multimodal~\cite{microsoft2025phi4}, and MiniCPM-o~\cite{yao2024minicpmv}, are of this kind, with a lineage traceable to multimodal architectures~\cite{chu2024qwen2audio,liu2023llava,alayrac2022flamingo,li2023blip2} and modality encoders~\cite{radford2021clip,zhai2023siglip,radford2022whisper,girdhar2023imagebind}. In the image processing literature, emotion and affect have long been read directly from visual signals, through visual emotion analysis and emotion distribution learning~\cite{yang2021solver,yang2021stimuli,yang2022seeking}, facial expression and micro-expression recognition~\cite{wang2020region,xia2020revealing,li2020joint}, and personality-aware image aesthetics~\cite{li2020personality}, while audio-visual correspondence modeling~\cite{min2020multimodal} and image captioning~\cite{ji2020spatio,liu2021vocabulary} connect visual signals to attention and text generation; omni-modal LLMs fold these capabilities into a single residual stream, which is exactly the setting we probe. In psychology, the dimensional organization of emotion is described by Russell's circumplex model~\cite{russell1980circumplex,posner2005circumplex} and the valence-arousal-dominance framework~\cite{mehrabian1996pad,mohammad2018vad}. A recent line of work examines emotion representation inside LLMs directly, characterizing the emotional latent space, localizing emotion-inference mechanisms, and discovering or controlling emotion circuits in text-only models~\cite{reichman2025emotions,tak2025mechanistic,wang2025emotioncircuits,dong2025rational}. Unlike this line, we use emotion as a controlled vehicle to reveal the reading--intervention layer separation and test its generality across three omni-modal architectures, moving past whether emotion representations exist or can be controlled within a single text modality.

\section{Method}

This section describes the experimental setup shared by \S4--\S6: models, data, activation processing and probing, the steering injection protocol, and evaluation. All experiments follow pre-registered decision thresholds, multi-seed bootstrap confidence intervals, and paired random-direction controls. Results that do not meet a pre-registered threshold are reported faithfully as findings rather than retried.

\subsection{Models}

We study three independently designed omni-modal LLMs. Qwen2.5-Omni-7B is the primary model. Its LLM backbone (based on Qwen2.5~\cite{yang2024qwen2}, the Thinker) has 28 layers and hidden dimension 3584, with hook path \textit{model.model.layers[0..27]}. For cross-model validation we use Phi-4-Multimodal-Instruct (Phi-4-Mini 3.8B LLM, 32 layers, hidden 3072, with a SigLIP-400M vision encoder, a 24-layer Conformer audio encoder, and a Mixture-of-LoRAs adapter) and MiniCPM-o-4.5 (36 layers, hidden 4096). The three models differ in LLM backbone, tokenizer, and modality-encoder lineage, so conclusions that hold across them are not easily attributed to shared components. All analyses are performed at inference time, with no model weights updated.

\subsection{Data}

The main analysis uses a self-constructed, fixed tri-modal sample set, M3: 5 emotions (anger / calm / fear / joy / sadness) $\times$ 3 modalities (text / audio / image) $\times$ 25 samples per class, for 375 samples in total (configuration in \textit{configs/multimodal\_m3\_samples.json}). A fixed sample set ensures comparability across modalities, layers, and seeds. Probe training and cross-modal transfer are carried out on M3.

To support external controls and scaled-up evaluation, we additionally use three public datasets: GoEmotions (text)~\cite{demszky2020goemotions}, RAVDESS speech-only (audio)~\cite{livingstone2018ravdess}, and EmoSet (image)~\cite{yang2023emoset}. The domain-shift control in the Supplement additionally uses two text emotion datasets, ISEAR~\cite{scherer1994isear} and DailyDialog~\cite{li2017dailydialog}. The non-text-context steering in \S6.2 is conditioned on 100 RAVDESS utterances and 102 EmoSet images, respectively.

\subsection{Activation Processing and Probing}

For each input, we extract the layer-$\ell$ hidden state with a forward hook and mean-pool over the token dimension to obtain a per-layer activation vector $h^{(\ell)} \in \mathbb{R}^{d}$. Probing is linear. We first fit a whitening transform $W_s$ on the training activations of source modality $s$ (per-feature standardization followed by decorrelation, which removes the domination of class centroids by shared principal axes), then train a linear emotion probe $f_s$ in the whitened space $\tilde{h} = W_s (h - \mu_s)$. Accuracy is reported with 5-fold cross-validation. Cross-modal transfer means applying the probe trained on source modality $s$ directly to classify activations of target modality $t$, with transfer accuracy defined as
\begin{equation}
\mathrm{Acc}_{s \to t}
= \frac{1}{|\mathcal{D}_t|} \sum_{(h, y) \in \mathcal{D}_t}
\mathbb{1}\!\left[\, f_s\!\left(W_s (h - \mu_s)\right) = y \,\right],
\label{eq:transfer}
\end{equation}
where $\mathcal{D}_t$ is the target-modality sample set and $y$ the emotion label. Whitening is necessary. In the raw space the largest one or two principal components dominate all class centroids, and naive nearest-class-centroid accuracy is only 0.24--0.40, and after whitening the same readout exceeds 0.90. Unless stated otherwise, statistics are reported as 95\% confidence intervals from 1000 bootstrap~\cite{efron1993bootstrap} resamples over 5 seeds.

\subsection{Steering Protocol}

Emotion directions are extracted independently at each injection layer $\ell$ (\textit{raw\_cmd\_pc\_orth\_k2}). On the layer-$\ell$ activations of M3 we take the means over target-emotion and neutral samples to obtain the contrastive activation mean difference $v_0^{(\ell)} = \bar{h}^{(\ell)}_{\text{emo}} - \bar{h}^{(\ell)}_{\text{neutral}}$, then orthogonalize away the top two dataset-level principal components $\{u^{(\ell)}_1, u^{(\ell)}_2\}$ of that layer to obtain the unit direction
\begin{equation}
v^{(\ell)} = \frac{v_0^{(\ell)} - \sum_{k=1}^{2} \big(v_0^{(\ell)\top} u^{(\ell)}_k\big)\, u^{(\ell)}_k}{\big\lVert v_0^{(\ell)} - \sum_{k=1}^{2} \big(v_0^{(\ell)\top} u^{(\ell)}_k\big)\, u^{(\ell)}_k \big\rVert},
\label{eq:direction}
\end{equation}
which reduces confounding with the shared principal axes. Injection uses a multi-layer normalized scheme. Over the injection-layer set $\mathcal{L}$ (L10/14/18/22 for Qwen), the layer-$\ell$ hidden state is updated as
\begin{equation}
h^{(\ell)} \leftarrow h^{(\ell)} + \alpha\, \lVert h^{(\ell)} \rVert \, v^{(\ell)},
\qquad \ell \in \mathcal{L},
\label{eq:injection}
\end{equation}
so that the norm of the injected vector at each layer is always $\lvert\alpha\rvert$ times the current hidden-state norm $\lVert h^{(\ell)} \rVert$ of that layer (per-layer normalized), making the perturbation magnitude comparable across layers. Cross-model experiments map $\mathcal{L}$ by normalized depth to L12/16/20/24 for Phi-4 and L14/18/22/27 for MiniCPM. The main experiments use $\alpha = +0.12$ (some experiments sweep over $\{\pm0.04, \pm0.08, \pm0.12\}$). Generation uses prefill-suffix + decode-all, with temperature 0.7, top-$p$ 0.9, and \textit{max\_new\_tokens} 80. Text steering uses 30 neutral prompts by default (scaled to 150 in \S6.2). Audio- and image-context steering use 100 RAVDESS utterances and 102 EmoSet images, respectively (\S6.2). Every condition is paired with a same-magnitude random-direction injection as control, and we monitor the degradation (OOD) rate. The single-layer control experiments in \S5.1--\S5.2 use the same direction-extraction and normalization protocol, varying only the number and position of injection layers.

\subsection{Evaluation}

Emotion intensity is read out by an external evaluator, decoupled from the subject model. The primary evaluator is \textit{j-hartmann/emotion-english-distilroberta-base}~\cite{hartmann2022emotionenglish} (hereafter VAD-BERT), whose native 7-emotion output we fold into the 5-emotion space of \S3.2 (anger / calm / fear / joy / sadness) as 5-class probabilities. Writing $p_c(x)$ for the evaluator's probability of target emotion $c$ on text $x$, the steering effect is defined as the difference between steered and baseline mean target-emotion probability over $N$ prompts,
\begin{equation}
\Delta_{\text{target}}
= \frac{1}{N} \sum_{i=1}^{N}
\Big[\, p_c\big(g_{\text{steer}}(x_i)\big) - p_c\big(g_{\text{base}}(x_i)\big) \,\Big],
\label{eq:delta}
\end{equation}
where $g_{\text{base}}$ and $g_{\text{steer}}$ are generation before and after injection. We additionally use the compound valence of NLTK VADER~\cite{hutto2014vader} as a cross-check on the valence dimension (its agreement with the VAD-BERT readout is reported as a Spearman correlation). To rule out potential coupling between the evaluator and the subject model, \S6.2 scales text steering to $n=150$ and introduces an independent GPT-5.5 as an external judge (LLM-as-a-judge~\cite{zheng2023llmjudge}), which outputs five-dimensional emotion probabilities and a fluency score. The per-sample agreement between GPT-5.5 and VAD-BERT is likewise reported as a Spearman correlation. The mechanistic analysis in \S5.4 uses the logit lens~\cite{nostalgebraist2020logitlens}, projecting each layer's hidden state directly through the unembedding matrix to the vocabulary to track the layer-wise rank of the target-emotion token. All steering experiments report the OOD rate. \S6.5 further verifies, via WikiText perplexity and GSM8K accuracy, that steering does not harm basic language or reasoning ability.

\section{Cross-Modal Emotion Geometry}

Before turning to the dissociation, we characterize the representation space the steering experiments operate on, the geometry of emotion across text, audio, and image inside Qwen2.5-Omni-7B. A linear emotion probe trained on one modality transfers above the random baseline (0.20) in all six cross-modal directions, confirming a non-trivial shared subspace. Transfer is asymmetric and strongest toward text (audio-to-text 0.558, image-to-text 0.568), so text carries the most readily readable component of the shared emotion information and we target text generation for steering in \S5; the aggregate text-target probe peaks at L27. A same-modality cross-domain control (ISEAR versus DailyDialog) rules out ordinary domain shift: the cross-domain drop of 16.49 percentage points is far smaller than the cross-modal decay, with a pure-modality increment of 26.09 points ($p = 0.0001$).

\begin{figure}[t]
\centering
\includegraphics[width=\columnwidth]{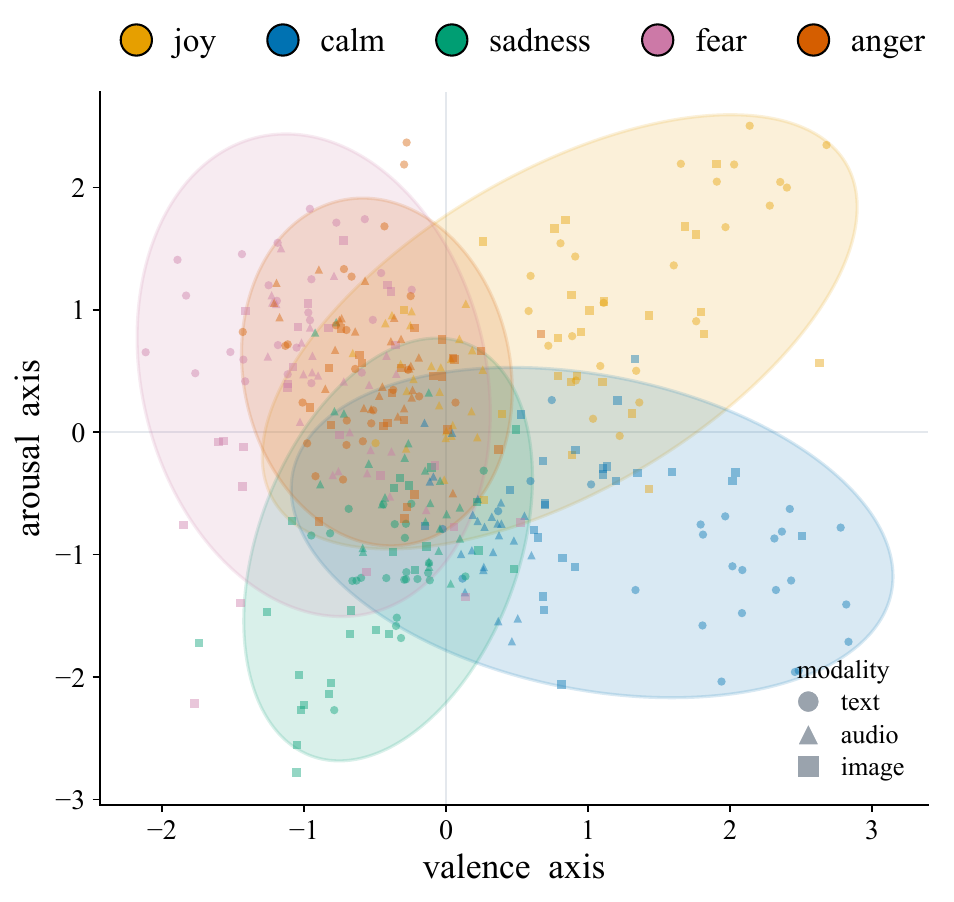}
\caption{Cross-modal emotion geometry in Qwen2.5-Omni-7B. Whitened M3 activations at L14 (375 samples) projected onto the supervised valence and arousal axes. Color encodes emotion, marker shape encodes modality, and shaded regions are 95\% confidence ellipses. The five emotions fall at their expected circumplex positions and the three modalities of each emotion overlap, visualizing the dimensional sharing reported in the Supplementary Material.}
\label{fig:emotion_geometry}
\end{figure}

The shared structure is organized by the two continuous valence--arousal dimensions rather than discrete categories. Single-axis valence and arousal transfer (0.853 and 0.696) far exceed 5-class transfer (0.558), and orthogonalizing the two axes collapses 5-class transfer to 0.176, near chance. \Cref{fig:emotion_geometry} visualizes this: the five emotions occupy their expected valence--arousal circumplex positions and the three modalities of each emotion overlap. This organization is Qwen-strong but not universal: it only partially replicates on Phi-4. The joy direction is the cross-model exception, occupying the unique high-valence, high-arousal corner of the circumplex~\cite{russell1980circumplex} and replicating robustly in both models, which anticipates joy as the strongest and most stable steering anchor in \S5--\S6. Full transfer tables, the depth anatomy, the domain-shift control, and the valence--arousal decoupling are reported in the Supplementary Material.

\section{Probing--Steering Dissociation}
\subsection{Dissociation in a Single Controlled Comparison}

A core assumption of representation engineering is that the layer at which a linear probe reads the strongest signal is also the layer at which activation steering is most effective. We test this assumption directly and under control.
On Qwen2.5-Omni-7B, the aggregate text-target best layer for cross-modal probe transfer is L27, the peak of the layer-wise mean of the audio-to-text and image-to-text transfer curves. We apply single-layer activation steering at this layer, holding the injection magnitude ($\alpha = +0.12$) and the direction-extraction protocol (\textit{raw\_cmd\_pc\_orth\_k2}) identical to the multi-layer configuration. The causal effect on generation is almost zero. The VAD-BERT external readout gives $\Delta_{\text{joy}} = +0.02$ with a 95\% bootstrap CI of $[-0.08, +0.13]$ that contains 0, and the paired random-direction control gives the same $\Delta_{\text{joy}} = +0.02$, so the main direction is essentially indistinguishable from a random one ($n = 30$ prompts, VADER cross-check $\Delta = 0.000$, OOD rate 0). Injecting the same emotion direction in a per-layer normalized way into the mid-to-late multi-layer combination (L10/14/18/22) tells a different story. It gives $\Delta_{\text{joy}} = +0.55$ with CI $[+0.39, +0.69]$, about 26 times the single-layer effect at L27 (full numbers in \Cref{tab:method_comparison}).

\begin{table}[t]
\centering
\caption{Four-condition joy steering on Qwen2.5-Omni-7B. All conditions use $\alpha=+0.12$, the same prompts ($n=30$) and direction protocol, differing only in layer selection. Ours and Baseline-B/C use the \textit{raw\_cmd\_pc\_orth\_k2} direction, Baseline-A the centroid. $\Delta_{\text{joy}}$ is the VAD-BERT change of steered relative to baseline (95\% CI from 1000 bootstrap), Random $\Delta$ is the same-condition random control, and the OOD rate is 0\% throughout. Visual comparison in \Cref{fig:method_comparison_bar}.}
\label{tab:method_comparison}
\resizebox{\columnwidth}{!}{
\begin{tabular}{lcccc}
\toprule
Condition & $\Delta_{\text{joy}}$ & 95\% CI & VADER $\Delta$ & Random $\Delta$ \\
\midrule
Ours (L10/14/18/22)
& +0.549
& $[+0.393, +0.693]$
& +0.201
& $-0.098$ \\

Baseline-A (centroid)
& +0.573
& $[+0.437, +0.707]$
& +0.242
& $-0.098$ \\

Baseline-B (L18)
& +0.152
& $[-0.004, +0.313]$
& +0.143
& $-0.067$ \\

Baseline-C (L27)
& \underline{+0.021}
& $[-0.078, +0.128]$
& 0.000
& +0.019 \\
\bottomrule
\end{tabular}
}
\end{table}

\begin{figure}[h]
\centering
\includegraphics[width=\linewidth]{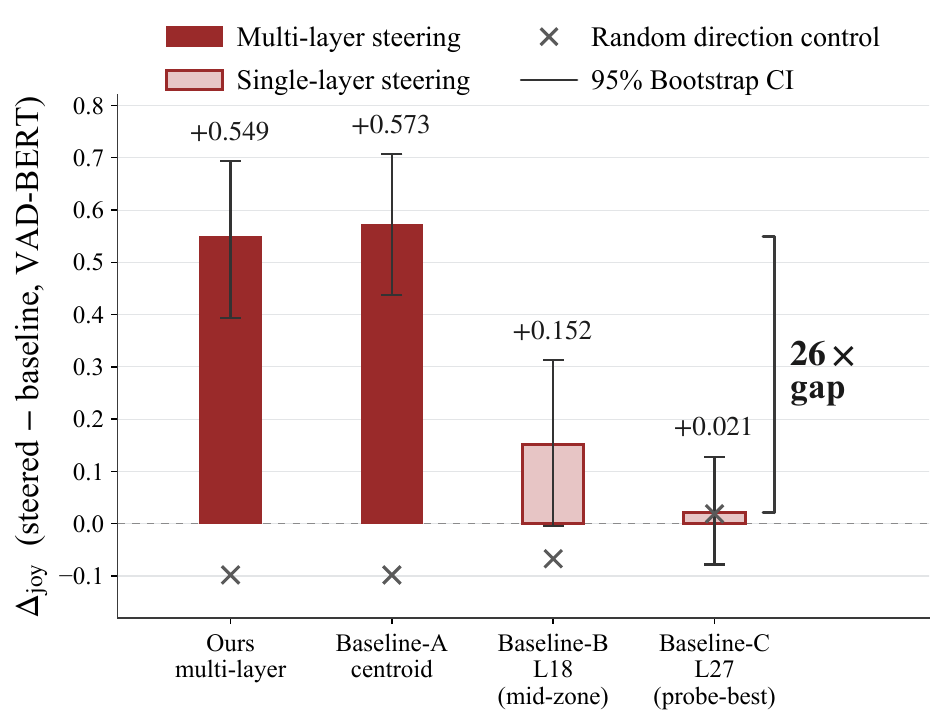}
\caption{Four-condition joy steering effect. Solid bars are $\Delta_{\text{joy}}$ of the main direction, lighter bars the same-condition random control, and error bars are 95\% bootstrap CIs. Multi-layer injection (Ours, Baseline-A) rises sharply, while single-layer probe-best L27 (Baseline-C) sits near zero and matches its random control. The roughly 26-fold gap is the empirical signature of the dissociation. Exact values in \Cref{tab:method_comparison}.}
\label{fig:method_comparison_bar}
\end{figure}

\Cref{fig:method_comparison_bar} makes the gap visually clear. The main-direction bar for Baseline-C (single-layer probe-best L27) is almost as low as its random-direction control bar, whereas the multi-layer bars (Ours / Baseline-A) rise sharply, and their 95\% bootstrap error bars do not overlap at all.

Layer position, not layer count, drives this gap. Baseline-B in \Cref{tab:method_comparison}, a \emph{single}-layer L18 injection in the mid-to-late range, already reaches $\Delta_{\text{joy}} = +0.15$, about 7 times the single-layer probe-best L27 ($+0.02$). With layer count held at one, moving the injection from L27 to the mid-zone L18 alone produces this 7-fold change, and the additional multi-layer gain (\S5.2) only stacks on top. The 26-fold gap is dominated by where one injects.

This \emph{readability--intervenability} gap is the empirical signature of the probing--steering dissociation: the layer where a linear probe reads most easily is not the layer where activation steering is most effective.

\subsection{Layer-Wise Causal Map}

\S5.1 established the dissociation through a single L27-vs-Ours comparison. We now give the full 28-layer causal map on Qwen2.5-Omni-7B and pre-register the dissociation criteria for the cross-model extension (\S5.3).

\begin{figure}[t]
\centering
\includegraphics[width=\linewidth]{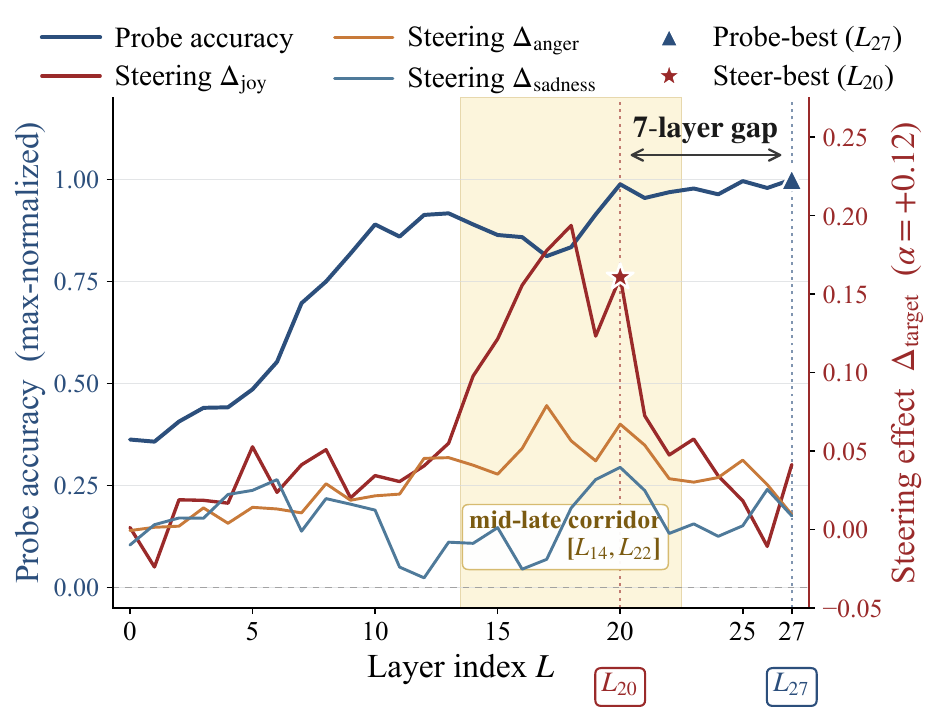}
\caption{The 28-layer dissociation curves of Qwen2.5-Omni-7B. Blue is layer-wise probing accuracy (left axis), and the colored lines are the single-layer steering effect for joy / anger / sadness (right axis, $\alpha=+0.12$). The probing peak (L27) and steering peak (L20) are 7 layers apart, and the shading marks the mid-to-late steering zone [L14, L22].}
\label{fig:qwen_layer_wise}
\end{figure}

\textbf{Single-layer steering sweep.}
We measure the single-layer activation steering effect independently for all 28 layers (3 emotions $\times$ 2 $\alpha$ $\times$ 5 seeds $\times$ 20 prompts, 16,800 generations in total). As shown in \Cref{fig:qwen_layer_wise} and \Cref{tab:qwen_single_layer}, the steering effect is near zero at both ends (L0--L5 and L25--L27) and peaks in the mid-to-late range. The single-layer best for joy is stable at L18 ($\alpha = +0.12$, $\Delta_{\text{joy}} = +0.194$), the best for anger is L17 ($\Delta_{\text{anger}} = +0.079$), and the best for sadness is L20 ($\Delta_{\text{sadness}} = +0.040$). Sadness is the least stable of the three, with a secondary peak at L5 under low $\alpha$.

\begin{table}[h]
\centering
\caption{Single-layer steering best layer and effect over the 28 layers of Qwen2.5-Omni-7B ($\alpha = +0.12$, $n = 20$ prompts $\times$ 5 seeds, VAD-BERT). All three emotions peak in the mid-to-late range (L17--L20), while the end layers are near 0. Multi-layer effect in \Cref{tab:method_comparison}.}
\label{tab:qwen_single_layer}
\begin{tabular*}{1\linewidth}{@{\extracolsep{\fill}}lcc}
\toprule
Emotion & Single-layer best & $\Delta_{\text{target}}$ \\
\midrule
joy     & L18 & +0.194 \\
anger   & L17 & +0.079 \\
sadness & L20 & +0.040 \\
\bottomrule
\end{tabular*}
\end{table}

\textbf{Pre-registered dissociation criteria.}
We define the dissociation as holding by either of two independent criteria: (i) the Pearson $r$ between layer-wise probing accuracy and steering effect over the 28 layers is $< 0.30$, or (ii) the absolute layer distance between the two curves' peaks is $\geq 5$ layers.

On the Qwen aggregate (text-target probing), the steering peak is L18 at $\alpha = +0.08$ and L20 at $\alpha = +0.12$, against a probing peak of L27 in both cases (a gap of 9 and 7 layers, with $r = 0.36$ and $0.45$, respectively). The peak-gap criterion passes at both values of $\alpha$. The Pearson $r$, though close, does not cross the 0.30 threshold, so we adopt the peak gap as the primary criterion and carry it over to the cross-model extension in \S5.3.

\textbf{Layer-pair synergy.}
After 12 selected layer-pair injection experiments, super-additivity appears in 8/12 pairs for joy and 7/12 for anger. These layer pairs cover combinations of shallow+shallow, shallow+final, middle+middle, middle+final, and final+final, and the strongest pairs consistently concentrate within the mid-to-late corridor, e.g., L14\_18, L18\_22, L14\_22. For sadness, 6/12 pairs are weaker than its strongest single layer. This shows that the mid-to-late steering zone holds at the single-layer level and is also the dense locus of pair-level synergy.

\Cref{fig:qwen_layer_wise} combines this evidence in one plot: the probing-accuracy curve (blue) climbs monotonically with depth and peaks at L27, while the three steering-effect curves (red/orange/blue) are single-peaked within the mid-to-late corridor and approach zero at both ends, with the two sets of peaks clearly offset. The shape of the curves alone displays the layer separation between reading and intervention.

Combining the single-layer curves, the peak gap, and pair synergy as three independent lines of evidence, the probing--steering dissociation on Qwen2.5-Omni-7B is confirmed at the level of a \emph{quantitative layer-wise causal map}. \S5.3 extends the same pre-registered pair of criteria to Phi-4-Multimodal and MiniCPM-o-4.5.

\subsection{Universality of the Steering-Effective Zone Across Three Models}

Does the dissociation quantified on Qwen2.5-Omni-7B (\S5.1, \S5.2) hold across architectures? We replicate the measurement under an identical protocol on Phi-4-Multimodal-Instruct (32 layers, hidden 3072) and MiniCPM-o-4.5 (36 layers, hidden 4096), which share no lineage with Qwen in LLM backbone, tokenizer, or modality encoder.

\begin{figure*}[h]
\centering
\includegraphics[width=\linewidth]{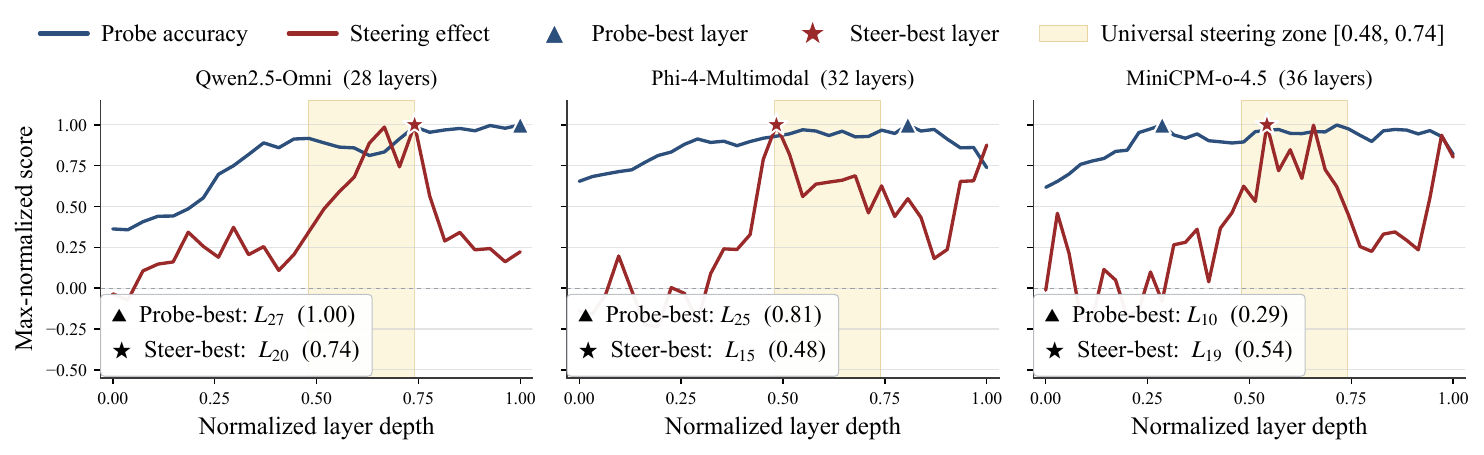}
\caption{Cross-architecture probing--steering dissociation (signature figure). Subplots are Qwen2.5-Omni, Phi-4-Multimodal, and MiniCPM-o-4.5. The probe-best layer (triangle) drifts across normalized depth 0.29--1.00, nearly the full network, while the steer-best layer (star) stays in the steering-effective zone [0.48, 0.74] (band) across all three architectures.}
\label{fig:three_model_dissociation}
\end{figure*}

The cross-model comparison reveals a striking asymmetry. The probe-best layer sits at normalized depths of 1.00 (Qwen L27), 0.81 (Phi-4 L25), and 0.29 (MiniCPM L10), spanning almost the full network depth. The steer-best layer, by contrast, stays in the mid-to-late residual stream, at normalized depths of 0.74 (Qwen L20), 0.48 (Phi-4 L15), and 0.54 (MiniCPM L19) (\Cref{tab:three_model_dissociation}, \Cref{fig:three_model_dissociation}). The absolute layer distances between probe peak and steer peak are 7, 10, and 9 layers, all exceeding the pre-registered dissociation threshold from \S5.2 ($\lvert\text{peak gap}\rvert \geq 5$ layers), and the layer-wise Pearson correlations are 0.45, 0.55, and 0.36, short of the \emph{highly correlated} range.

The steer-best layer of all three architectures lands in the same narrow band of normalized depth $[0.48, 0.74]$, while the probe-best layer scatters across $[0.29, 1.00]$, nearly the entire network. This contrast holds across the three models with no counterexample: the steering-effective zone is stable across architectures, the probe-best layer is not. MiniCPM shows an order reversal, its probe-best layer sitting in the early network (L10) ahead of its steer-best (L19), opposite to Qwen and Phi-4. The reversal sits entirely on the volatile probing side. MiniCPM's steer-best (0.54) still falls inside the universal band. Why MiniCPM's probing--steering ordering differs is a question of mechanism, which we take up in \S5.5 and \S7.2.

\begin{table}[t]
\centering
\caption{Cross-architecture probing--steering dissociation across three models (Qwen2.5-Omni, Phi-4-Multimodal, MiniCPM-o-4.5). Probe and steer peaks are measured per model by a layer-wise scan (5 seeds, VAD-BERT); each cell gives the layer and its normalized depth (layer / $L_{\max}$). Under the pre-registered criterion ($\lvert\text{peak gap}\rvert \geq 5$ layers), the gap passes in all three (7 / 10 / 9 layers), and the steering peak (0.48--0.74) is far more concentrated than the probing peak (0.29--1.00).}
\label{tab:three_model_dissociation}
\resizebox{\columnwidth}{!}{
\begin{tabular}{lccccc}
\toprule
Model & Layers & Probe-best & Steer-best & Gap & $r$ \\
\midrule
Qwen      & 28 & L27 (1.00) & L20 (0.74) & 7  & +0.452 \\
Phi-4     & 32 & L25 (0.81) & L15 (0.48) & 10 & +0.546 \\
MiniCPM   & 36 & L10 (0.29) & L19 (0.54) & 9  & +0.362 \\
\bottomrule
\end{tabular}
}
\end{table}

The three subplots of \Cref{fig:three_model_dissociation} make this invariance immediate: the steer-best stars of all three models fall into the same pale-yellow band (normalized depth [0.48, 0.74]), while the probe-best triangles scatter across 0.29, 0.81, and 1.00. This is the strongest cross-architecture evidence that reading and intervention are functionally distinct operations on the residual stream.

\subsection{A Logit-Lens Mechanism}

\S5.1--\S5.3 quantified the empirical strength of the probing--steering dissociation, but \emph{why} it holds remains open. We use the logit lens~\cite{nostalgebraist2020logitlens} to measure, after projecting each layer's hidden state directly through the unembedding matrix, the timing at which the target-emotion token enters the top-50 vocabulary candidates.

On Qwen2.5-Omni-7B ($n = 30$ prompts $\times$ 3 emotions $= 90$ trajectories, emotion-conditioned, with the logit lens run as a deterministic forward pass with no sampling-seed dimension), the median layer at which the target emotion token first enters the top-50 vocabulary candidates is L23 (the vocabulary-commitment layer). Combined with the probing-saturation layer established in \S5.2 (L20, where the linear probe reaches 95\% of its peak) and the steering-peak layer (L19, the aggregate peak of the \S5.2 layer-wise sweep), we obtain a clear three-stage ordering, with the steering peak at L19 first, then probing saturation at L20, and finally vocabulary commitment at L23.

The three pre-registered hypotheses are adjudicated in \Cref{tab:logit_lens_hypotheses}. H2 (the steering-effective layer is adjacent to the commitment window, pre-registered threshold $\lvert\text{gap}\rvert \leq 4$) holds, with $\lvert \text{L19} - \text{L23} \rvert = 4$ layers, exactly at the threshold. H3 (probing saturates before vocabulary commitment) also holds, with L20 before L23. H1 (commitment precedes saturation) does not hold: saturation is 3 layers earlier than commitment.

\begin{table}[t]
\centering
\caption{Adjudication of the three pre-registered logit-lens hypotheses (Qwen2.5-Omni-7B, 90 emotion-conditioned trajectories; layer numbers are medians of the timing).}
\label{tab:logit_lens_hypotheses}
\begin{tabularx}{\columnwidth}{c X c}
\toprule
Hypothesis & Statement & Verdict \\
\midrule
H1 & Vocabulary commitment precedes probing saturation (L23 $<$ L20)
& $\times$ \\
H2 & The steering peak is adjacent to the commitment window ($\lvert \text{L19} - \text{L23} \rvert \leq 4$)
& $\checkmark$ \\
H3 & Probing saturation precedes vocabulary commitment (L20 $<$ L23)
& $\checkmark$ \\
\bottomrule
\end{tabularx}
\end{table}

\begin{figure}[t]
\centering
\includegraphics[width=\linewidth]{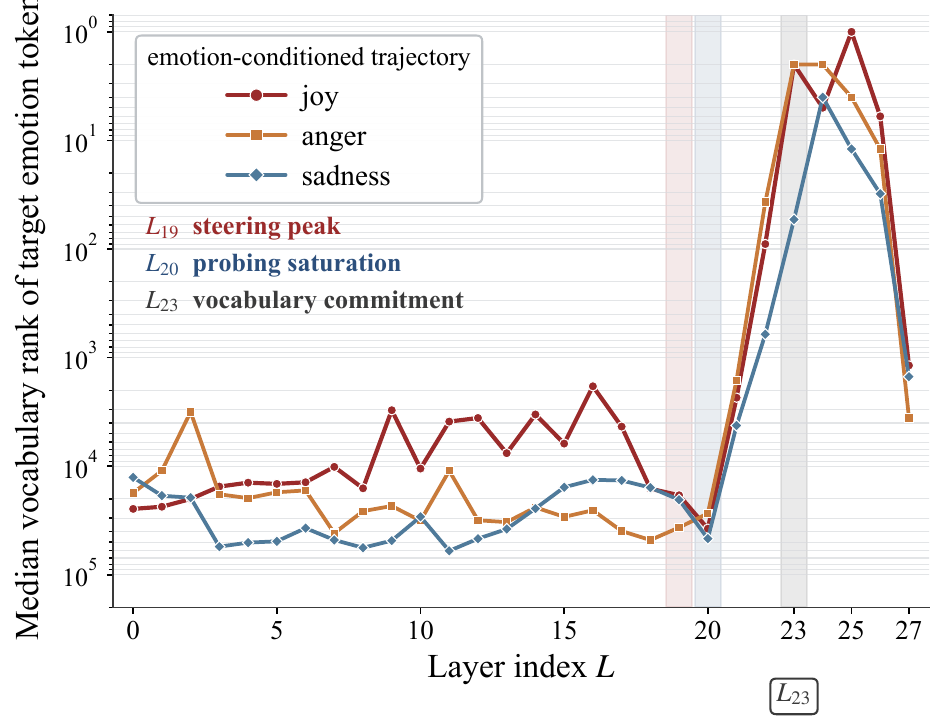}
\caption{Cross-layer logit-lens trajectory of the emotion token (joy / anger / sadness, median rank over $n=30$ prompts). The vertical axis is the median vocabulary rank of the target emotion token (log scale, inverted, so the top is rank 1). Shaded bands mark the three landmark layers: steering peak (L19), probing saturation (L20), and vocabulary commitment (L23).}
\label{fig:logit_lens_trajectory}
\end{figure}

\Cref{fig:logit_lens_trajectory} shows the layer-wise rank trajectory of the three emotion tokens: over L0--L22 the median rank of the target emotion token stays at the $10^4$ scale (buried deep inside the vocabulary) and only at L23 jumps sharply to the top of the vocabulary (rank $\leq 2$). This abrupt change confirms the timing of vocabulary commitment and shows that at the layers before commitment (including steering peak L19 and probing saturation L20) the emotion information is already encoded but not yet linearized into a token choice.

The fact that H1 \emph{does not hold} is itself informative. It indicates that the representation stabilizes first and is then linearized into token logits. The causal window for activation steering thus lies in the stable-but-uncommitted intermediate state, near L19. The probe-best layer L27, already well past the L23 commitment, is where the model has committed to its token choice, so an injected perturbation can no longer rewrite the already-linearized logit structure. This timing mechanistically explains the phenomenon in \S5.1 whereby single-layer injection at L27 collapses to a level indistinguishable from a random direction.

\subsection{A Predictive Functional Model}

The three-stage ordering in \S5.4 suggests a specific functional form. The effectiveness of steering at layer $L$ should depend on two factors at once: (i) the information at that layer is already sufficiently encoded, i.e., the probe accuracy exceeds a threshold $\theta$, and (ii) there is still enough downstream amplification room after that layer, i.e., the number of remaining layers exceeds a threshold $\tau$.
We write this hypothesis as a two-factor model:
\[
\begin{aligned}
\mathrm{steer}(L)
&\propto
\max\!\left(0,\, \mathrm{probe}(L) - \theta\right) \\
&\quad \times
\max\!\left(0,\, L_{\max} - L - \tau\right).
\end{aligned}
\]

\begin{figure*}[t]
\centering
\includegraphics[width=\textwidth]{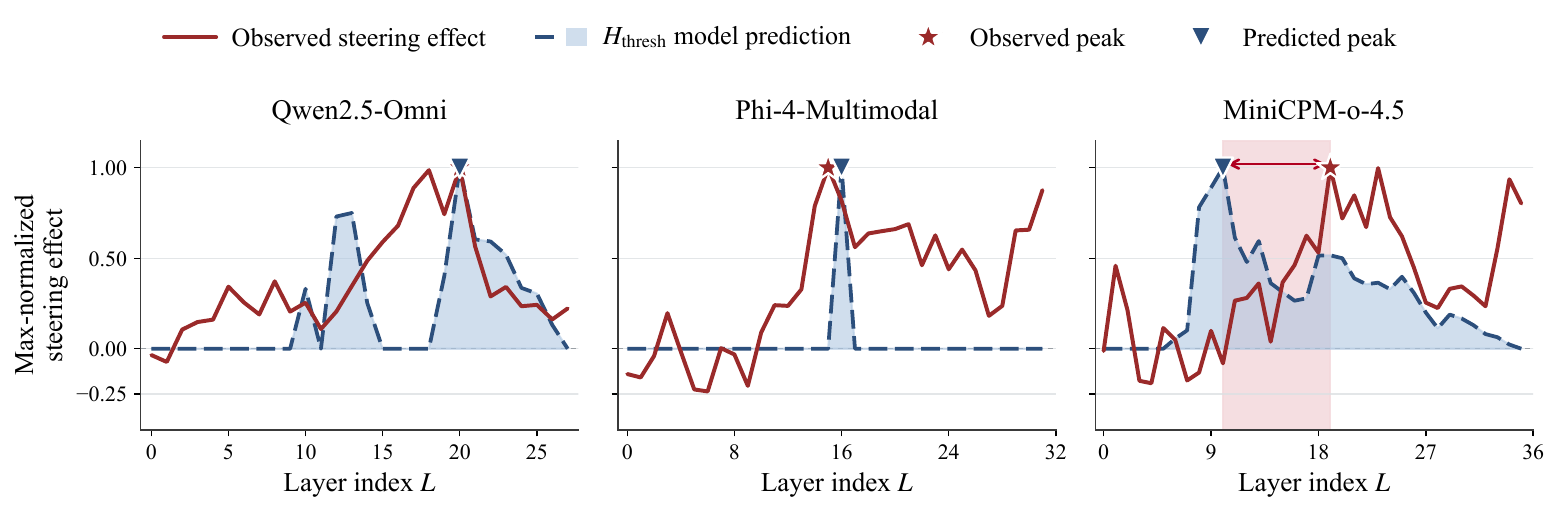}
\caption{$H_{\mathrm{thresh}}$ prediction versus the measured steering effect across models. Solid lines are measured, dashed lines the $H_{\mathrm{thresh}}$ prediction, stars the measured peak, and triangles the predicted peak. Qwen is hit exactly (error 0), Phi-4 is off by 1 layer, and MiniCPM by 9 layers (red shading marks the gap). Fit parameters in the Supplementary Material.}
\label{fig:hthresh_fit}
\end{figure*}

We grid-search $(\theta, \tau)$ independently for each model to predict the steering peak layer (\Cref{fig:hthresh_fit}). On Qwen2.5-Omni, $(\theta = 0.54, \tau = 0)$ predicts a steering peak of L20, exactly matching the measured L20. On Phi-4-Multimodal, $(\theta = 0.24, \tau = 14)$ predicts L16 against a measured L15, an error of 1 layer or about 3\% of network depth. On MiniCPM-o-4.5, $(\theta = 0.24, \tau = 0)$ predicts L10 but the measured peak is L19, an error of 9 layers and a clear failure (\Cref{fig:hthresh_fit}).

The exact/approximate success on Qwen and Phi-4 quantitatively supports the causal reading of the \emph{stable-but-uncommitted} timing in \S5.4, where steering effectiveness $\approx$ representational readability $\times$ downstream plasticity. But MiniCPM's failure is equally structural. MiniCPM's probe-best layer is in the early network (L10), and $H_{\mathrm{thresh}}$ predicts that the steer peak should likewise be early, yet the measured peak is at L19, 9 layers later than predicted. This deviation cannot be explained by the two factors of \emph{information encoding $\times$ downstream room,} suggesting a third architecture-related factor, a model-specific lower bound of the actionable zone, which dictates that even when early-layer information is already encoded, steering takes effect only once the representation enters this intervenable band.

This residual deviation is the most concrete open problem for a mechanistic theory of steering. Does a cross-architecture universal actionable zone exist? If so, how can it be measured directly, independent of the two diagnostics of probing and the logit lens? We take this up as a central topic of the discussion and future work (\S7.3).

\section{Cross-Modal and Cross-Model Causal Control}

Can multi-layer injection in the mid-to-late zone located by the dissociation causally control the emotion of generation across three input modalities and three models? We test this, validate it with an independent LLM judge, characterize the cross-model specificity of joy, and confirm that steering preserves the model's basic abilities.

\subsection{Multi-Layer Normalized Injection Protocol}

Building on the dissociation result of \S5, we use multi-layer normalized injection across the mid-to-late effective zone rather than a single probe-best layer: the \textit{raw\_cmd\_pc\_orth\_k2} direction is injected at L10/14/18/22 of Qwen, per-layer normalized, with $\alpha = +0.12$ (full protocol in \S3.4). Emotion intensity is read out by the external VAD-BERT evaluator with a same-condition random-direction paired control, and the OOD rate is monitored throughout.

\subsection{Tri-Modal Controllability}

\begin{figure*}[t]
\centering
\includegraphics[width=\linewidth]{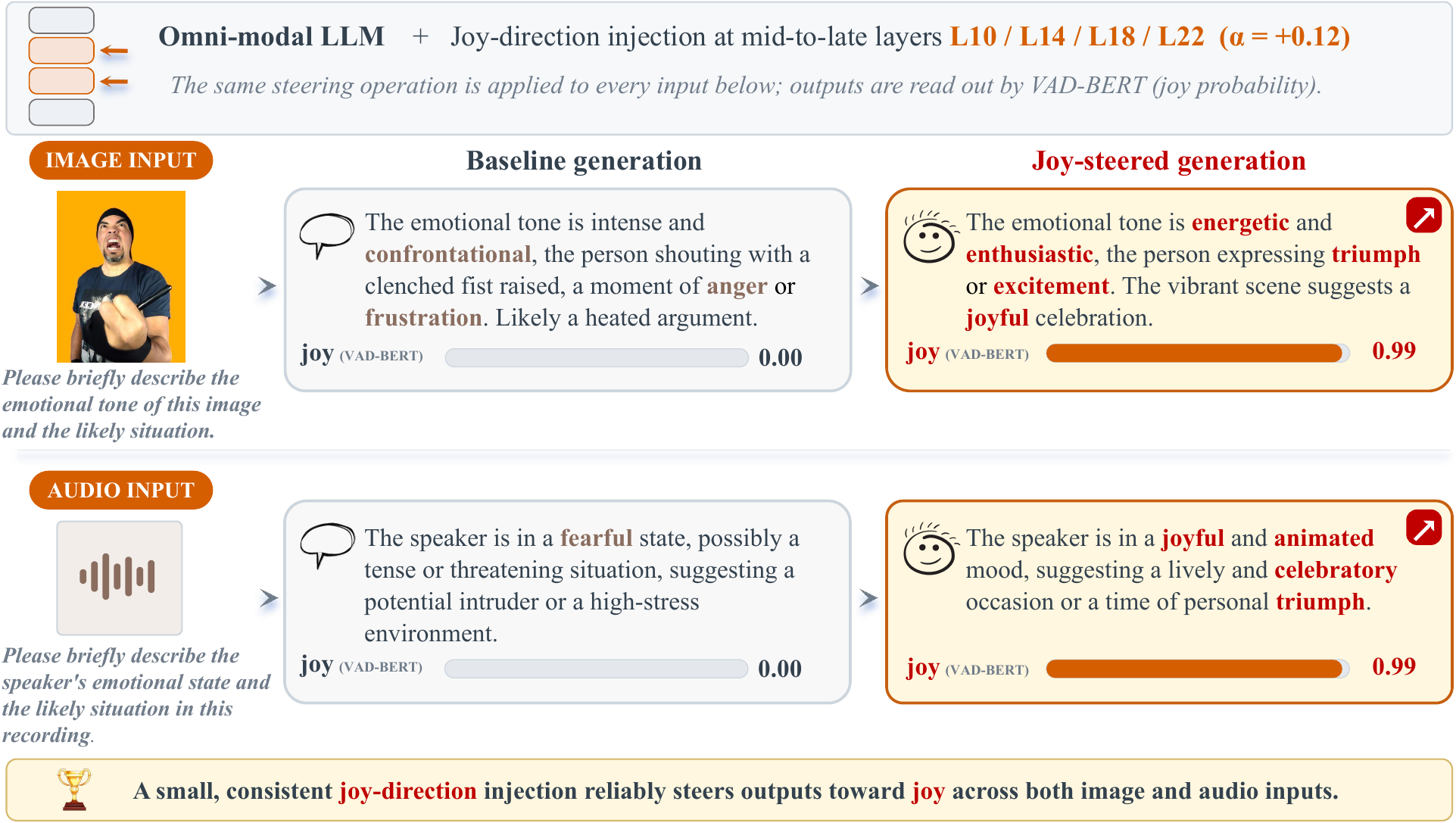}
\caption{Qualitative cross-modal joy steering. The same multi-layer joy injection (L10/14/18/22, $\alpha = +0.12$) is applied to an image input (EmoSet, anger) and an audio input (RAVDESS, sad speech). The baseline generation conveys the original negative tone, while the joy-steered generation shifts to a celebratory one, with the VAD-BERT joy probability rising from $0.00$ to $0.99$ in both cases. Quantitative results in \Cref{tab:three_modality_steering}.}
\label{fig:qualitative_steering}
\end{figure*}

We test the same protocol under three input contexts: text-only (30 prompts), audio-conditioned (100 RAVDESS utterances), and image-conditioned (102 EmoSet images). \Cref{tab:three_modality_steering} gives the VAD-BERT $\Delta_{\text{target}}$ at $\alpha = +0.12$ for the joy / sadness / anger directions under the three modalities. Across the nine cells of three modalities and three emotions, joy is strong under all three modalities ($+0.33$ to $+0.67$), while sadness / anger show moderate, context-dependent positive effects. The OOD rate is 0 in all conditions, and paired random-direction controls are near zero. This shows that the multi-layer effective zone located in \S5 remains causally effective on text generation and in non-text omni-modal contexts. \Cref{fig:qualitative_steering} illustrates this qualitatively for an image and an audio input: the same injection turns a negative baseline description into a joyful one in both cases, with the external VAD-BERT joy probability rising from $0.00$ to $0.99$.

\begin{table}[h]
\centering
\caption{Multi-layer emotion steering under three input contexts (VAD-BERT $\Delta_{\text{target}}$, $\alpha=+0.12$; text 30 prompts, audio 100 RAVDESS, image 102 EmoSet). joy is strongly positive under all three modalities, the OOD rate is 0, and random controls are near zero.}
\label{tab:three_modality_steering}
\begin{tabular}{lccc}
\toprule
Input context & $\Delta_{\text{joy}}$ & $\Delta_{\text{sadness}}$ & $\Delta_{\text{anger}}$ \\
\midrule
text-only        & +0.549 & +0.101 & +0.166 \\
audio-context    & \textbf{+0.670} & +0.361 & +0.282 \\
image-context    & +0.334 & +0.162 & +0.179 \\
\bottomrule
\end{tabular}
\end{table}

\textbf{Independent LLM-judge validation.}
VAD-BERT is an automatic evaluator homologous to the subject model. To rule out evaluator coupling, we scale text steering to $n=150$ prompts and rescore with an independent GPT-5.5 as the external judge. The joy direction gives $\Delta_{\text{joy}} = +0.246$ under GPT-5.5 (95\% CI $[+0.211, +0.283]$, paired random control $-0.007$), and the per-sample Spearman correlation between GPT-5.5 and VAD-BERT reaches $\rho = 0.809$. This magnitude is comparable to the automatic-versus-human agreement reported by the human-evaluation study of style-vector steering~\cite{stylevectors2026humaneval} (mean $r = 0.776$ across over 7{,}000 crowdsourced ratings, ICC $= 0.71$--$0.87$), supporting the automatic evaluator as a credible proxy for emotion intensity and confirming that the joy effect is not an artifact of a single evaluator. Sadness ($+0.023$) and anger ($+0.024$) show only weak effects under GPT-5.5, and anger's VAD-BERT--GPT correlation is low ($\rho = 0.19$), consistent with the dimensional structure of \S4. Joy occupies the unique position of the circumplex and has the cleanest effect, while mixed-valence emotions have weak effects and low evaluator agreement.

\subsection{Cross-Model Steering Replication}

We replicate joy steering on Phi-4-Multimodal and MiniCPM-o-4.5 under the same G15 protocol (layer positions mapped to each model by normalized depth). \Cref{tab:cross_model_steering} shows that Phi-4 replicates a strong effect ($\Delta_{\text{joy}} = +0.415$, CI excluding 0, all pre-registered criteria passed), about 25\% weaker than Qwen, consistent with the model-size difference (3.8B vs 7B). MiniCPM-o gives a null result ($\Delta_{\text{joy}} = +0.022$, CI containing 0). MiniCPM's failure is not random. It corroborates the failure of $H_{\mathrm{thresh}}$ to predict MiniCPM in \S5.5 (peak error 9 layers) and the unstable geometric diagnostics of MiniCPM in \S4, jointly marking MiniCPM as a boundary-condition model rather than a simple replication failure.

\begin{table}[h]
\centering
\caption{Cross-model replication of joy steering (VAD-BERT $\Delta_{\text{joy}}$, $\alpha=+0.12$, layers mapped by normalized depth). Qwen and Phi-4 replicate a strong effect (CI excludes 0); MiniCPM-o is null (CI contains 0), matching its boundary-condition diagnostics in \S4 / \S5.5.}
\label{tab:cross_model_steering}
\begin{tabular}{lccc}
\toprule
Model & Injection layers & $\Delta_{\text{joy}}$ & 95\% CI \\
\midrule
Qwen2.5-Omni      & L10/14/18/22 & \textbf{+0.549} & $[+0.393, +0.693]$ \\
Phi-4-Multimodal  & L12/16/20/24 & \textbf{+0.415} & $[+0.289, +0.536]$ \\
MiniCPM-o-4.5     & L14/18/22/27 & +0.022 & $[-0.064, +0.122]$ \\
\bottomrule
\end{tabular}
\end{table}

\subsection{Joy as a Cross-Model Anchor}

Why is joy the most stable across modalities and models? We test the directional specificity of each emotion with a $5\times5$ cross-injection matrix (GPT-5.5 judge), scoring each direction by its diagonal increment minus the largest off-diagonal increment. Joy is the only emotion specific (above the 0.15 threshold) in both Qwen and Phi-4 (0.363 and 0.469); calm passes only on Phi-4, and the mixed-valence emotions pass in neither. Joy occupies the unique high-valence, high-arousal corner of the circumplex, so its direction overlaps least with neighbors, giving the strongest and most consistent specificity. This joy-anchored separability is a cross-model invariant, consistent with the dimensional organization of \S4. The full specificity matrix is in the Supplementary Material.

\subsection{Preservation of Fluency and Reasoning Ability}

Steering changes emotion without harming basic abilities. Joy steering ($\alpha=+0.12$) raises WikiText~\cite{merity2016wikitext} perplexity only moderately (5.62 to 8.39, far below the PPL $>50$ of destructive steering), leaves the GPT-5.5 fluency score essentially unchanged ($\Delta_{\text{fluency}} = -0.03$), and does not lower GSM8K~\cite{cobbe2021gsm8k} accuracy (16\% to 18\%, within noise). Further detail is in the Supplementary Material.

\section{Discussion}

\subsection{Implications for Representation Engineering}

The results of \S5 refute the standard assumption that the layer where a probe reads a concept best is also where intervention works best. On Qwen2.5-Omni, single-layer injection at the probe-best layer is indistinguishable from a random direction while mid-to-late multi-layer injection is about 26 times stronger (\Cref{tab:method_comparison}), and the gap holds across all three architectures, with the probe-best layer spanning nearly the full network depth and the steer-best layer confined to a narrow mid-to-late band (\Cref{tab:three_model_dissociation}).

This yields a directly practical corollary. Selecting the injection layer by probing accuracy is a mistaken heuristic. Under the common practice of choosing the layer at the probing peak, it can place intervention at causally inert layers and thus severely underestimate a method's true controllability. By contrast, the mid-to-late steering-effective zone (normalized depth about $0.5$--$0.75$) is stable across architectures and provides a layer-selection criterion that does not depend on any single model's probing curve. The robust gain of multi-layer normalized injection over single-layer injection (\S5.2) further shows that spreading the injection within this zone is more reliable than betting on a single layer. The boundary of this criterion is set by MiniCPM. Although its steering-effective zone still lies in the mid-to-late range, $H_{\mathrm{thresh}}$ fails to predict its peak (\Cref{fig:hthresh_fit}), indicating that the criterion gives an effective \emph{range} rather than a single point that can be extrapolated exactly, and that precise cross-architecture localization still requires model-specific calibration.

\subsection{The Geometric Basis of the Dissociation}

Why does the dissociation hold at the functional level? The logit-lens ordering of \S5.4 (steering peak, then probing saturation, then vocabulary commitment) places \emph{reading} and \emph{intervention} at different stages of the forward pass. A probe reads the strongest signal where the representation \emph{saturates} and is fully linearly separable, whereas steering is most effective in the earlier window where the representation is \emph{stable but not yet committed to a token}. Readability measures whether the information \emph{is already present}. Intervenability measures whether it \emph{can still be rewritten}. The two are functionally distinct operations on the residual stream, with no a priori reason to share the same layer.

The $H_{\mathrm{thresh}}$ two-factor model of \S5.5 makes this falsifiable: steering effectiveness scales with the product of information already encoded and downstream amplification room, hitting the peak exactly on Qwen and within one layer on Phi-4 (\Cref{fig:hthresh_fit}). MiniCPM exposes what the two factors miss: its steering peaks late despite early-encoded information with ample downstream room, pointing to a model-specific lower bound on the actionable zone that the representation must enter before intervention takes hold. The gap is in the predictive formula, not the empirical finding, as MiniCPM's steer-best layer still lies in the cross-architecture mid-to-late band. The dissociation is thus general in substance and architecture-dependent in form: the steering-effective zone is stable across architectures, while the exact inter-layer geometry, especially where the probing peak lands, is set by the architecture and remains to be formally characterized (\S7.3).

\subsection{Future Work}

This work leaves several directly extensible directions. The first is the \emph{actionable zone} open problem raised in \S5.5. Whether a cross-architecture universal intervenable band exists, and whether it can be measured directly and independently of the probing and logit-lens diagnostics, is the key to explaining the MiniCPM deviation and completing the third factor of $H_{\mathrm{thresh}}$. Second, the dissociation framework can in principle transfer to other intervention tasks that require layer selection. Our preliminary exploration shows that on a safety steering task the mid-to-late layers located by the dissociation outperform the probe-best layer, suggesting that the framework has applicability beyond emotion generation. The directions used in that exploration come from safety contrastive activations, not the emotion subspace, and their cross-model stability is unresolved, so we treat it as directional evidence of the framework's transferability and leave it to future systematic study.

\section{Limitations}

This work has several boundaries to state explicitly. First, all three models are at the 7B scale, and whether the general mid-to-late steering-effective zone is preserved as scale extends to 70B+ has not been verified. Second, the emotion labels and datasets are mainly in English, and whether the emotion structure replicates across languages and cultures remains open. Third, we use linear probes and linear directions throughout, and non-linear readouts (such as a sparse autoencoder) might reveal structure that linear methods miss, which this work does not cover. Fourth, causal validation concentrates on injection toward text generation, and within-modality steering (e.g., injecting an audio direction to produce audio output) is left to future work. Fifth, $H_{\mathrm{thresh}}$ is exact on Qwen, approximate on Phi-4, and fails on MiniCPM (\Cref{fig:hthresh_fit}), indicating that the two-factor model does not yet capture some architecture-related actionable-zone factor, and we have listed its formal characterization as a central open problem (\S7.3). Each of these boundaries points to a specific extension, not a confounder for the core conclusion. The dissociation itself holds robustly across three independent architectures and does not depend on relaxing any of the above conditions.

\section{Conclusion}

Reading and intervention are not the same operation on the residual stream of omni-modal language models. The layer where a linear probe reads a concept most strongly and the layer where activation steering is most effective are functionally separate: on Qwen2.5-Omni the causal gap is roughly 26-fold (\Cref{tab:method_comparison}), and across the three independent architectures the probe-best layer scatters across nearly the full network depth while the steer-best layer converges in a narrow mid-to-late band (\Cref{tab:three_model_dissociation}). A logit-lens account (causal handle, then probing saturation, then vocabulary commitment) and the falsifiable $H_{\mathrm{thresh}}$ model locate the steering window in the stable-but-uncommitted stage, instantiated in a cross-modal emotion geometry organized by valence--arousal and anchored across models by joy (\S4). The long-assumed \emph{readable implies steerable} premise of representation engineering thus does not hold in omni-LLMs: the mid-to-late zone is the locus of cross-architecture-stable causal intervention, while the probe-best layer only reflects representational saturation.

\section*{Acknowledgments}
This work was supported in part by the National Natural Science
Foundation of China (Grant No.~62227807 and Grant No.~U24B20186),
in part by the Brain Science and Brain-like Intelligence
Technology---National Science and Technology Major Project
(No.~2021ZD0200600, No.~2021ZD0200408),
and in part by the WQ \& UCAS Research Academy Intelligent
Computing Center (WRA-ICC) and the Supercomputing Center of
Lanzhou University.

\bibliographystyle{IEEEtran}
\bibliography{references}

\section{Biography Section}
\vspace{-4em}
\newcommand{\bioimg}[1]{\includegraphics[width=0.75in,height=0.94in,clip,keepaspectratio]{#1}}
\newcommand{\biogap}{\vspace*{-3em}}

\begin{IEEEbiography}[{\includegraphics[width=1in,height=1.25in,clip,keepaspectratio]{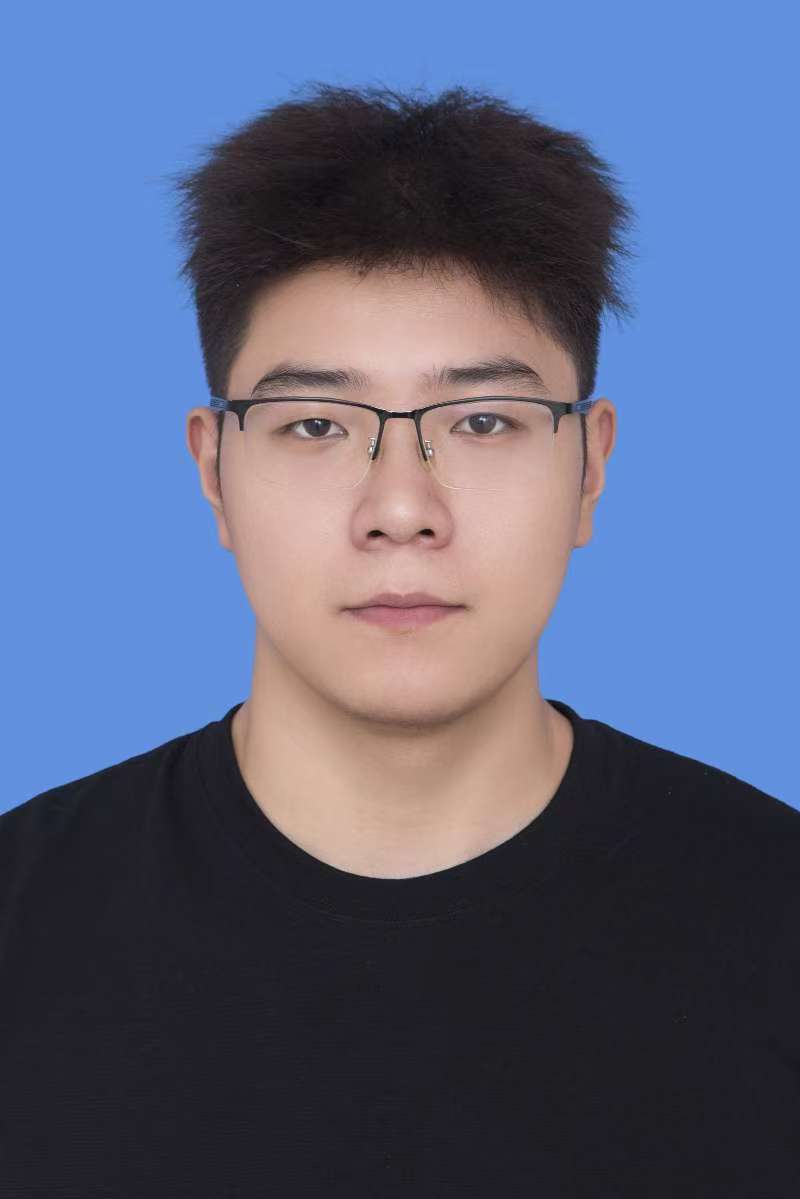}}]{Yibo Wang}
is currently pursuing the M.S. degree at the Gansu Provincial Key Laboratory of Wearable Computing, School of Information Science and Engineering, Lanzhou University, Lanzhou, China. He received the B.S. degree from Dalian University of Technology, Dalian, China. His research interests include multimodal large language models, affective computing, and reinforcement learning.
\end{IEEEbiography}
\biogap

\begin{IEEEbiography}[{\includegraphics[width=1in,height=1.25in,clip,keepaspectratio]{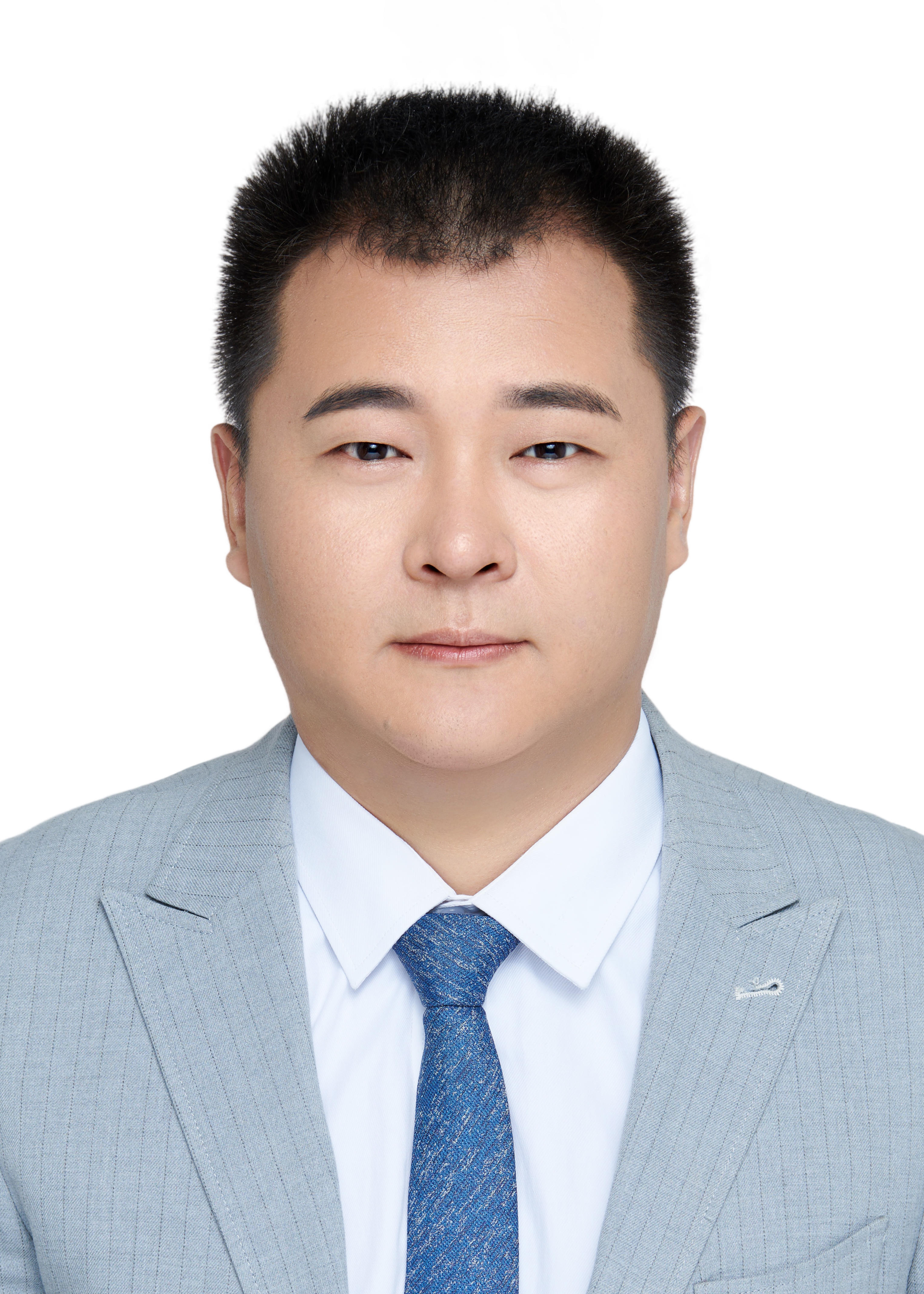}}]{Jisheng Dang}
received the Ph.D. degree from Sun Yat-sen University, China, advised by Prof. Jianhuang Lai and Prof. Huicheng Zheng. He worked as a research fellow at the NExT++ laboratory of the National University of Singapore, advised by Prof. Tat-Seng Chua. He is now a tenured associate professor at the School of Information Science and Engineering, Lanzhou University. His research interests include multimodal learning, video understanding, and embodied intelligence. He has published several papers as the first author or corresponding author in major journals and conferences including IEEE TPAMI/TIP/TNNLS/TITS/IJCAI/AAAI/ICLR/NeurIPS/PR.
\end{IEEEbiography}
\biogap

\begin{IEEEbiography}[{\includegraphics[width=1in,height=1.25in,clip,keepaspectratio]{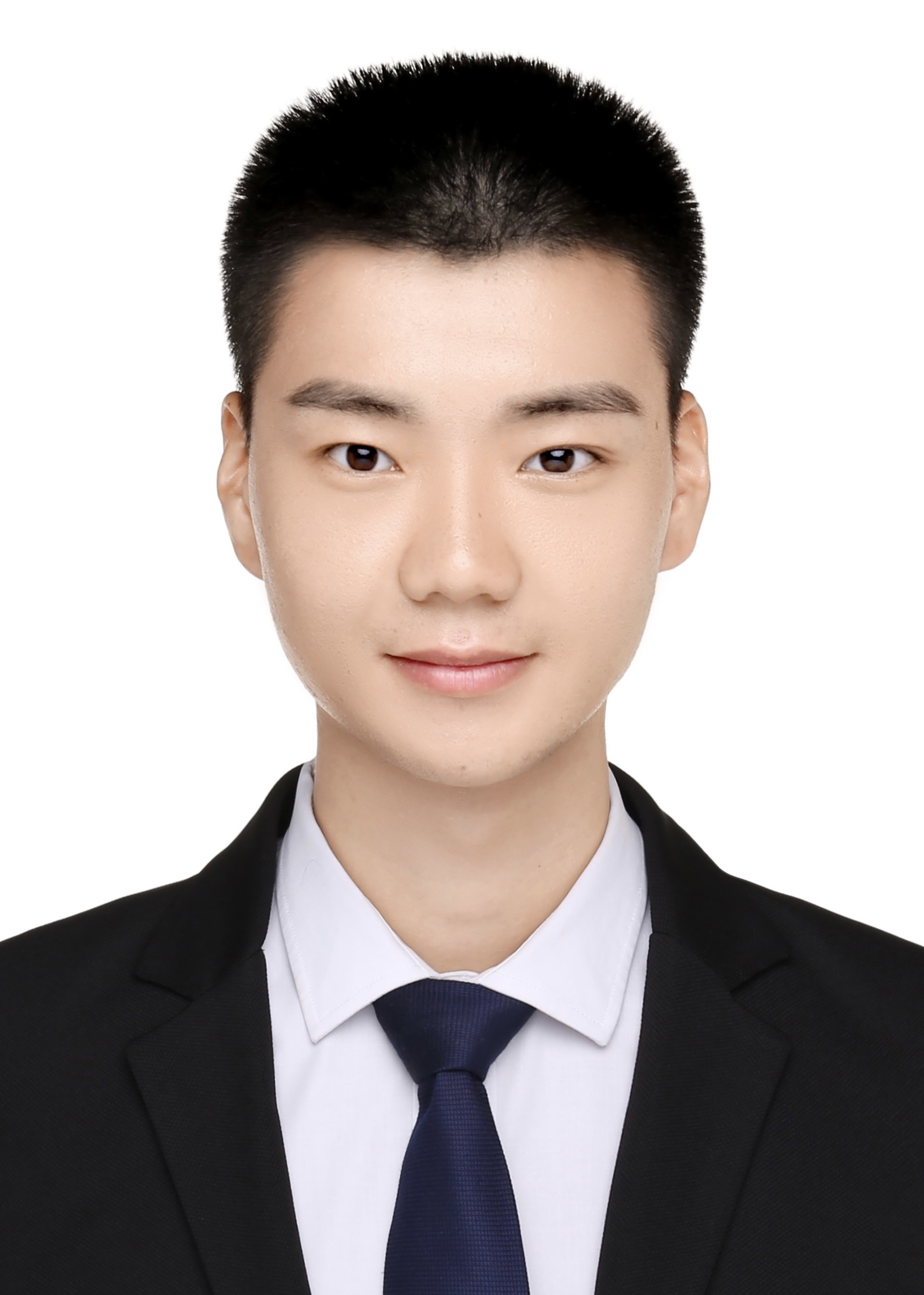}}]{Yitao Wu}
received the B.S. degree in information system and information management from Hainan University. His research interests include large language models, vision-language-action models, and vision-language models.
\end{IEEEbiography}

\begin{IEEEbiography}[{\includegraphics[width=1in,height=1.25in,clip,keepaspectratio]{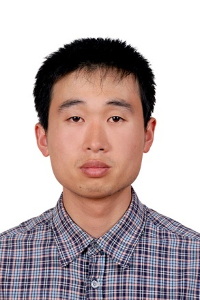}}]{Hong Peng}
received the Ph.D. degree from Lanzhou University, Lanzhou,
China. From 2010 to 2011, he was a Visiting Scholar with the
Institute of Computer System, ETH Zurich, Switzerland. He is
currently an Associate Professor with the School of Information
Science and Engineering, Lanzhou University. He is also in
charge of three projects from the National Natural Science
Foundation of China, the Central College Foundation Project of
Lanzhou University, and the Youth Cross-Project of Lanzhou
University. He has authored or coauthored more than 30 papers
in peer-reviewed journals, conferences, and book chapters.
His research areas include bioinformation processing and
ubiquitous affective computing.
\end{IEEEbiography}
\biogap

\begin{IEEEbiography}[{\includegraphics[width=1in,height=1.25in,clip,keepaspectratio]{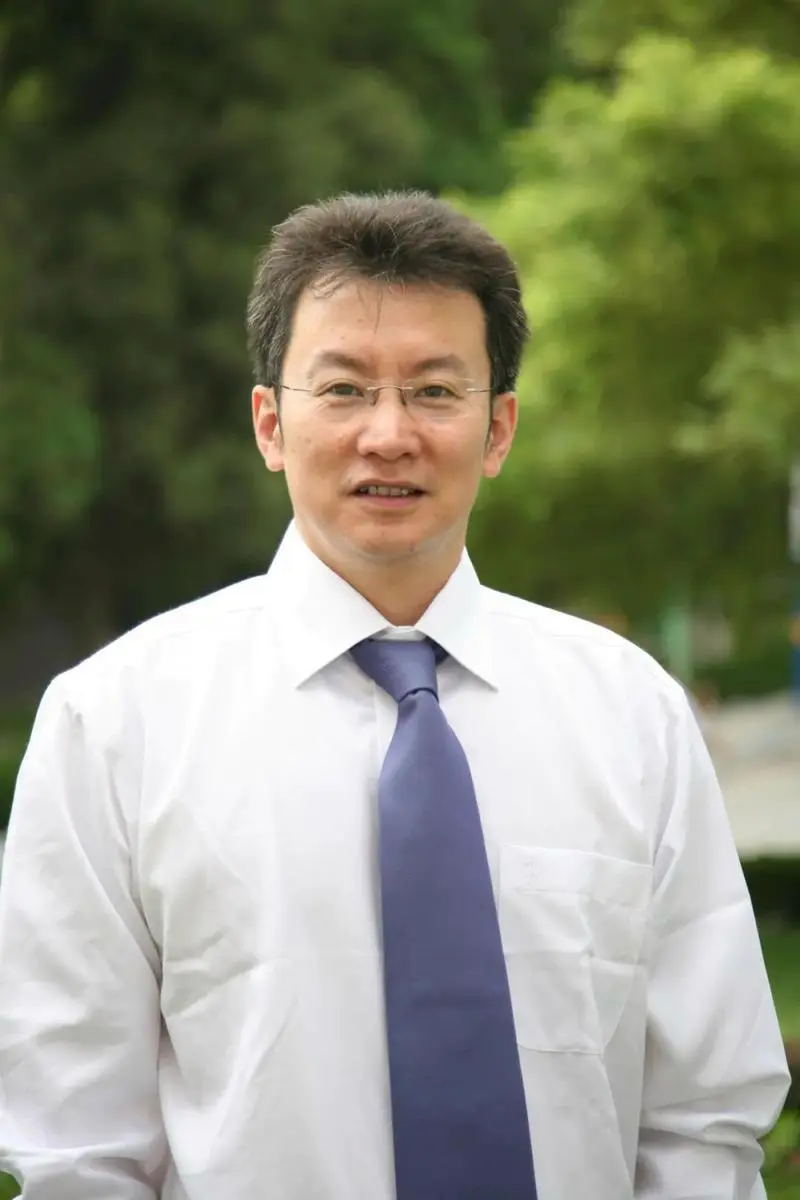}}]{Bin Hu (Fellow, IEEE)}
received the Ph.D. degree in computer science from the Institute of Computing Technology, Chinese Academy of Sciences, China, in 1998. Since 2008, he has been a Professor and Dean of the School of Information Science and Engineering, Lanzhou University. He has also held a guest professorship at ETH Zurich. He serves as Editor-in-Chief of IEEE Transactions on Computational Social Systems and is a Fellow of IET and AAIA. His research interests include pervasive computing, computational psychophysiology, data modeling, and artificial intelligence.
\end{IEEEbiography}
\biogap

\begin{IEEEbiography}[{\includegraphics[width=1in,height=1.25in,clip,keepaspectratio]{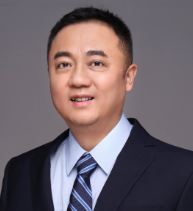}}]{Qi Tian (Fellow, IEEE)}
received the Ph.D. degree in ECE from the University of Illinois at Urbana-Champaign, Champaign, IL, USA, in 2002. He is currently the Chief Scientist of Huawei Terminal BG. He was the Chief Scientist in computer vision with Huawei Noah's Ark Laboratory from 2018 to 2020. Before he joined Huawei, he was a Full Professor with the Department of Computer Science, The University of Texas at San Antonio, San Antonio, TX, USA, from 2002 to 2019. He was listed in the top ten of the 2016 Most Influential Scholars in Multimedia by Aminer.org. He is an IEEE Fellow (Class 2016) and an Academician of the International Eurasian Academy of Sciences (elected in 2021), with 113,457 citations on Google Scholar.
\end{IEEEbiography}

\begin{IEEEbiography}[{\includegraphics[width=1in,height=1.25in,clip,keepaspectratio]{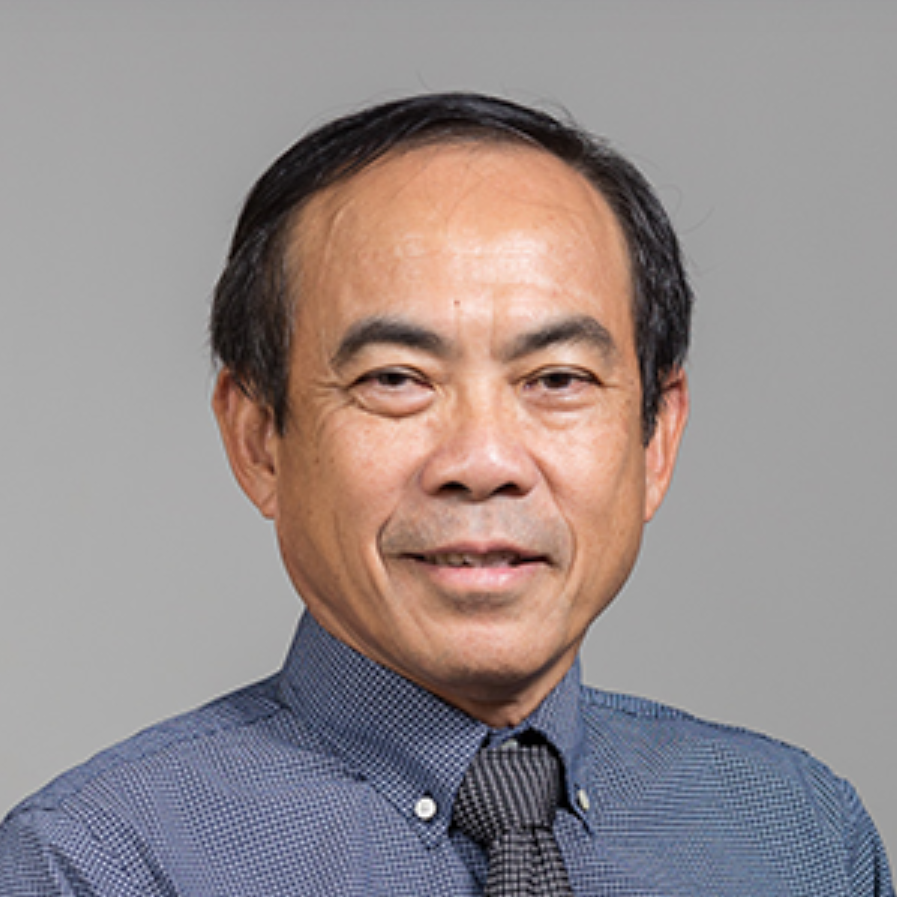}}]{Tat-Seng Chua}
received the Ph.D. degree from the University of Leeds, U.K. He is the KITHCT chair professor with the School of Computing, National University of Singapore, where he was the acting and founding dean of the School from 1998 to 2000. He is the co-director of NExT, a joint center between NUS and Tsinghua University, to develop technologies for live social media search. He is the 2015 winner of the prestigious ACM SIGMM Award. He is the chair of the Steering Committee of the ACM International Conference on Multimedia Retrieval (ICMR) and the Multimedia Modeling (MMM) conference series. He is also the general co-chair of ACM Multimedia 2005, ACM CIVR (now ACM ICMR) 2005, ACM SIGIR 2008, and ACM Web Science 2015. He serves on the editorial boards of four international journals. He is the co-founder of two technology startups in Singapore and a Fellow of the Singapore Academy of Sciences, with 114,163 citations on Google Scholar.
\end{IEEEbiography}
\biogap

\vfill

\end{document}